\documentclass{article}

\usepackage{PRIMEarxiv}

\usepackage[utf8]{inputenc} %
\usepackage[T1]{fontenc}    %
\usepackage{hyperref}       %
\usepackage{url}            %
\usepackage{booktabs}       %
\usepackage{amsfonts}       %
\usepackage{nicefrac}       %
\usepackage{microtype}      %
\usepackage{lipsum}
\usepackage{fancyhdr}       %
\usepackage{graphicx}       %
\graphicspath{{media/}}     %

\title{Toward Efficient and Incremental Spectral Clustering via Parametric Spectral Clustering 
}

\author{
  Jo-Chun Chen, Hung-Hsuan Chen \\
  Computer Science and Information Engineering \\
  National Central University \\
  Taoyuan, Taiwan\\
  \texttt{logchen0620@g.ncu.edu.tw, hhchen1105@acm.org} \\
}

\usepackage[utf8]{inputenc} %
\usepackage[T1]{fontenc}    %
\usepackage{hyperref}       %
\usepackage{url}            %
\usepackage{booktabs}       %
\usepackage{amsfonts}       %
\usepackage{nicefrac}       %
\usepackage{microtype}      %
\usepackage{xcolor}         %
\usepackage{algorithm}
\usepackage[noend]{algpseudocode}
\usepackage{tikz}
\usetikzlibrary{shapes.multipart, positioning}
\usepackage{amsmath}
\usepackage{bm}
\usepackage{color}
\usepackage{graphicx}
\usepackage{makecell}
\usepackage{multirow}
\usepackage{subfigure}
\usepackage{tabularx}
\usepackage{mathtools}
\usepackage{pythonhighlight}
\usepackage{listings}
\usepackage{graphicx}
\usepackage{float}
\usepackage{subfigure} 
\usepackage{pdfpages}
\usepackage{comment}

\definecolor{pink}{rgb}{0.858, 0.188, 0.478}
\usepackage{xspace}

\definecolor{commentcolor}{RGB}{110,154,155}   %

\begin{document}
\maketitle

\begin{abstract}

Spectral clustering is a popular method for effectively clustering nonlinearly separable data. However, computational limitations, memory requirements, and the inability to perform incremental learning challenge its widespread application. To overcome these limitations, this paper introduces a novel approach called parametric spectral clustering (PSC). By extending the capabilities of spectral clustering, PSC addresses the challenges associated with big data and real-time scenarios and enables efficient incremental clustering with new data points. Experimental evaluations conducted on various open datasets demonstrate the superiority of PSC in terms of computational efficiency while achieving clustering quality mostly comparable to standard spectral clustering. The proposed approach has significant potential for incremental and real-time data analysis applications, facilitating timely and accurate clustering in dynamic and evolving datasets. The findings of this research contribute to the advancement of clustering techniques and open new avenues for efficient and effective data analysis. We publish the experimental code at \url{https://github.com/109502518/PSC_BigData}.

\end{abstract}


\section{Introduction} 

Clustering, as a fundamental technique in various fields such as data mining and pattern recognition, plays a critical role in analyzing large datasets by dividing them into smaller and more coherent groups, facilitating the analysis of large datasets.  The importance of clustering is further underscored by its frequent usage as a preprocessing step for subsequent analyses~\cite{lien2019visited}. Despite the extensive development of clustering techniques, it remains an active research area due to its wide-ranging applications and inherent challenges~\cite{kleinberg2003impossibility}.

Among the various clustering methods, spectral clustering has gained considerable popularity and demonstrated superior performance. Spectral clustering has a significant benefit in that it can manage datasets with patterns that are not linearly separable. This is accomplished by dividing the data points in a transformed space rather than directly dividing them in the original feature space. This unique property makes spectral clustering suitable for a variety of complex applications and domains, and it is often able to uncover clusters that other clustering techniques are unable to detect.

Although spectral clustering is a widely used clustering method that has shown remarkable performance in various applications, it suffers from significant limitations: its computation time and memory usage are huge, and it cannot perform incremental clustering. Therefore, it could be challenging to apply spectral clustering to a large dataset.  Furthermore, once a spectral clustering model is trained, it cannot cluster new data points without retraining the entire model. These limitations can be critical issues in applications with big data or when the data streams continuously arrive, and the clustering results must be updated in real-time.

To overcome these limitations and enable spectral clustering for handling big data and supporting incremental clustering, this paper presents an innovative approach called parametric spectral clustering (PSC). Essentially, PSC learns a parametric function to project data points into low-dimensional representations. This is a critical step in spectral clustering, although it requires a lot of computing resources. The approximated low-dimensional representations of both the new and original data points are then utilized altogether for clustering.

We evaluate the effectiveness and efficiency of PSC based on real-world datasets.  The experimental results demonstrate that PSC handles big data more efficiently and enables incremental clustering while maintaining clustering quality comparable to the standard spectral clustering method. Thus, PSC has broad applications in various fields with large datasets, such as user behavior segmentation, social media content grouping, real-time anomaly detection, and many more~\cite{wang23detecting, bai2019co, wu2023detecting, lin2021learning}.

The rest of the paper is organized as follows. Section~\ref{sec:rel-work} reviews previous studies on spectral clustering and incremental clustering. Section~\ref{sec:method} introduces the training process, the inference process, and the properties of PSC. Section~\ref{sec:exp} presents the experimental results.  Finally, Section~\ref{sec:disc} summarizes our contributions and discusses future research directions.
\section{Related Work}\label{sec:rel-work}

Spectral clustering leverages the eigenvalues and eigenvectors of a Laplacian matrix to transform the data into a lower-dimensional but more representative space, where a base clustering algorithm, such as $k$-means, can be applied to the transformed data instances. Spectral clustering has been extensively discussed in the literature, with notable works including~\cite{ng2001spectral,von2007tutorial}. Numerous variants and enhancements of spectral clustering have been proposed to improve its robustness and scalability~\cite{shi2000normalized, zhu2018low, couillet2016kernel, chen2010parallel}.

Incremental clustering (a.k.a. online clustering) addresses the challenge of clustering large datasets in real-time. Unlike traditional batch clustering, incremental clustering algorithms continuously cluster new data points (and possibly update the cluster ID of the original data points) as the new data points arrive. This incremental property allows efficient handling of dynamic datasets and enables the detection of evolving patterns.  Various approaches have been proposed for incremental clustering, including stream clustering algorithms~\cite{ding2015research, aggarwal2018survey} and online clustering methods~\cite{barbakh2008online}. Famous algorithms alone this line includes CluStream~\cite{aggarwal2003framework}, DenStream~\cite{cao2006density}, and BIRCH~\cite{zhang1996birch}.  However, these models are mostly used in the data stream environment and are sometimes sensitive to the selected hyperparameters.  Additionally, studies on data streams commonly assume that old data points become less important, which may ease the computation burden and accelerate the computation.  However, gradually retiring old data points may not be a valid assumption in many cases.  A tutorial on online clustering can be found in~\cite{montiel2022online}.

Previous studies have addressed the computational challenges associated with spectral clustering, primarily through approximation or sampling techniques~\cite{yan2009fast, li2011time}. In contrast to these methods, our proposed method is distinct by leveraging the parametric modeling technique, making PSC stand out in comparison to prior work.
\section{Methodology} \label{sec:method}

\subsection{Preliminary: Spectral Clustering}

\begin{algorithm}[tb]
\caption{Spectral clustering algorithm} \label{alg:sc}
\begin{algorithmic}[1]
\Require{$\boldsymbol{X}=\{\boldsymbol{x}_1, \ldots, \boldsymbol{x}_n\}$: $n$ data points, each $\boldsymbol{x}_i \in R^{d}$}
\Require{$p$: the length of the low-dimensional space, $p<d$}
\Ensure{$\boldsymbol{C} = \{c_1, \ldots, c_n$\}: the cluster IDs for the input data points $\boldsymbol{x}_1, \ldots, \boldsymbol{x}_n$}
      \State Compute the similarity matrix $\boldsymbol{S} = [s_{i,j}] \in R^{n \times n}$, where $s_{i,j}$ represents the similarity score between $\boldsymbol{x}_i$ and $\boldsymbol{x}_j$
      \State Compute the normalized Laplacian matrix $\boldsymbol{L} = \boldsymbol{D}^{-1/2}\boldsymbol{S}\boldsymbol{D}^{-1/2}$, where $\boldsymbol{D} = [d_{i,j}] \in R^{n \times n}$ is a diagonal matrix with $d_{i,i}= \sum_{j=1}^n s_{i,j}$ 
      \State Compute the $p$ eigenvectors $\boldsymbol{v}_1, \ldots, \boldsymbol{v}_p$ of $\boldsymbol{L}$ whose corresponding eigenvalues are the $p$ largest ones and form the matrix $\boldsymbol{V} = [\boldsymbol{v}_1, \ldots, \boldsymbol{v}_p] \in R^{n \times p}$ (i.e., the columns of $\boldsymbol{V}$ are the eigenvectors)
      \State Treat each row in $\boldsymbol{V}$ as a data point in $R^p$ and apply a clustering algorithm (e.g., $k$-means) on these data points.  The output cluster IDs $c_1, \ldots c_n$ are that of the data points $\boldsymbol{x}_1, \ldots \boldsymbol{x}_n$.
\end{algorithmic}
\end{algorithm}

Spectral clustering transforms input data points into a low-dimensional (but probably more representative) space and clusters the data points in the low-dimensional space. The transformation uses spectral properties, often identifying distinctive patterns not easily detected by alternative clustering algorithms.

The spectral clustering algorithm separates data points according to several key steps~\cite{ng2001spectral}. The algorithm begins by computing a similarity matrix (typically using the Gaussian kernel as a proxy of the similarity score, which will be explained in Section~\ref{sec:psc-train}) that captures the pairwise distances between the input data points. This similarity matrix is the foundation for constructing a graph representation of the dataset, with the data points acting as nodes, and the edges and weights reflecting the pairwise similarities. The algorithm calculates the Laplacian matrix from this graph representation that captures the relationships among the data points. The next step involves calculating the eigenvalues and eigenvectors of the Laplacian matrix. Finally, these eigenvectors serve as the transformed lower-dimensional representation, where a base clustering algorithm, such as $k$-means, can be applied. Algorithm~\ref{alg:sc} provides the pseudocode.

Although spectral clustering is powerful, it requires many computational resources in terms of time and space. First, the computational demands associated with spectral clustering are intensive. Tasks such as eigenvalue decomposition (as seen in line 3 of Algorithm~\ref{alg:sc}) and matrix operations (as seen in line 2 of Algorithm~\ref{alg:sc}) are computationally expensive and typically exhibit a time complexity of $O(n^3)$. Furthermore, the memory requirements for storing the similarity matrix and performing eigenvalue decomposition can quickly become prohibitive as the size of the dataset increases with the complexity of the space of $O(n^2)$~\cite{li2011time}. These computational and memory constraints limit the scalability of spectral clustering.

Another limitation of spectral clustering is the lack of support for incremental clustering. Once the spectral clustering model is trained on a fixed dataset, incorporating new data points, even just a single data point, into the existing model without retraining the entire model becomes challenging. This limitation hinders the application of spectral clustering in scenarios involving enormous datasets clustering or data stream clustering, as it becomes impractical to retrain the model whenever new data points arrive in real-time.

\subsection{Parametric Spectral Clustering -- Training} \label{sec:psc-train}

\begin{algorithm}[tb]
\caption{Training procedure of PSC} \label{alg:psc-train}
\begin{algorithmic}[1]
\Require{$\boldsymbol{X}=\{\boldsymbol{x}_1, \ldots, \boldsymbol{x}_n\}$: $n$ data points, each $\boldsymbol{x}_i \in R^{d}$}
\Require{$p$: the length of the low-dimensional space, $p<d$}
\Require{$r$: sampling rate, $0 < r \le 1$}
\Ensure{A trained model $M$}
\State $\nu = \textrm{round}(rn)$
\State Sample $\nu$ data points $\boldsymbol{X}'=\{\boldsymbol{x}_1', \ldots, \boldsymbol{x}_{\nu}'\}$ from $\boldsymbol{X}$
\State Compute the similarity matrix $\boldsymbol{S} = [s_{i,j}] \in R^{\nu \times \nu}$, where $s_{i,j}$ represents the similarity score between $\boldsymbol{x}_i'$ and $\boldsymbol{x}_j'$ (usually using Gaussian kernel)
\State Compute the normalized Laplacian matrix $\boldsymbol{L} = \boldsymbol{D}^{-1/2}\boldsymbol{S}\boldsymbol{D}^{-1/2}$, where $\boldsymbol{D} = [d_{i,j}] \in R^{\nu \times \nu}$ is a diagonal matrix with $d_{i,i}= \sum_{j=1}^n s_{i,j}$ 
\State Compute the $p$ eigenvectors $\boldsymbol{v}_1, \ldots, \boldsymbol{v}_p$ of $\boldsymbol{L}$ whose corresponding eigenvalues are the $p$ largest ones and form the matrix $\boldsymbol{V} = [\boldsymbol{v}_1, \ldots, \boldsymbol{v}_p] \in R^{\nu \times p}$, i.e., with the eigenvectors as columns of $\boldsymbol{V}$
\State Train a neural network $M$ (structure in Table~\ref{tab:nn}) that maps $\boldsymbol{x}_i$ to $\boldsymbol{v}_{i*}$ (row $i$ of matrix $\boldsymbol{V}$)
\end{algorithmic}
\end{algorithm}

\begin{table}[tb]
\centering
\caption{The structure of a simple neural network that learns to map $\boldsymbol{x}_i \in R^{d}$ to a smaller vector $\boldsymbol{v}_{i*} \in R^{p}~(p<d)$.}
\label{tab:nn}
\begin{tabular}{@{}ccc@{}}
\toprule
Layer Type                   & Number of Neurons & Activation Function \\ \midrule
Input feature                & $d$                 & None                \\
First fully connected layer  & $n_1$                & ReLU                \\
Second fully connected layer & $n_2$                & ReLU                \\
Third fully connected layer  & $n_3$                & ReLU                \\
Fourth fully connected layer & $p$                 & None                \\ \bottomrule
\end{tabular}
\end{table}

Parametric spectral clustering consists of two key steps. The first step is almost identical to the first half of the spectral clustering algorithm. Given a sampling rate $r~(0 < r \le 1)$, we sample $\nu = \textrm{round}(nr)$ data points $\boldsymbol{X}' = \{\boldsymbol{x}_1', \ldots, \boldsymbol{x}_{\nu}'\}$ from the initially available data points $\boldsymbol{X} = \{\boldsymbol{x}_1, \ldots, \boldsymbol{x}_n\}$.  We treat each data point $\boldsymbol{x}_i' \in R^d$ as a node and connect all edges in the graph. Next, we select a kernel function (typically a Gaussian kernel~\cite{ng2001spectral}) to define $s_{i,j}$ the similarity between two data points $\boldsymbol{x}_i'$ and $\boldsymbol{x}_j'$. Once the similarity matrix $\boldsymbol{S} = \left[s_{i,j}\right] \in R^{\nu \times \nu}$ is generated, we construct the Laplacian matrix $\boldsymbol{L}$ and compute $\boldsymbol{v}_1, \ldots, \boldsymbol{v}_p$ the $p$ eigenvectors of $\boldsymbol{L}$ whose eigenvalues are the largest.  We create a matrix $\boldsymbol{V} = [v_{i,j}] \in R^{\nu \times p}$ where the $i$th row in matrix $\boldsymbol{V}$ is denoted as $\boldsymbol{v}_{i*}$.  This matrix $\boldsymbol{V}$ is a crucial component in the subsequent training phase.

The second step aims to train a supervised learning model $M$ that maps a training instance $\boldsymbol{x}_i$ to $\boldsymbol{v}_{i*}$, the $i$th row of matrix $\boldsymbol{V}$, in which the columns are the $p$ largest eigenvectors of the Laplacian matrix $\boldsymbol{L}$. In other words, the model $M$ learns to project a data instance onto the directions of the $p$ most important principal components. We apply a simple multilayer perceptron (MLP) for the model $M$, whose structure is illustrated in Table~\ref{tab:nn}. However, it can be easily replaced with other neural networks or supervised learning models.

The training procedure for the parametric spectral clustering algorithm is given in Algorithm~\ref{alg:psc-train}.

\subsection{Parametric Spectral Clustering -- Inference}

\begin{algorithm}[tb]
\caption{Inference (clustering) procedure of PSC} \label{alg:psc-inf}
\begin{algorithmic}[1]
\Require{$\boldsymbol{X}=\{\boldsymbol{x}_1, \ldots, \boldsymbol{x}_n, \boldsymbol{x}_{n+1}, \ldots, \boldsymbol{x}_{n+m}\}$: $n+m$ data points.  The first $n$ data points are available initially; the extra $m$ data points need to be clustered but available after model training}
\Require{A trained model $M$}
\Ensure{$\boldsymbol{C} = \{ \boldsymbol{c}_1, \ldots, \boldsymbol{c}_n, \boldsymbol{c}_{n+1}, \ldots, \boldsymbol{c}_{n+m}\}$: the cluste IDs of $n+m$ data points in $\boldsymbol{X}$}
  \For{$i = 1$ to $n+m$}
    \State $\boldsymbol{u}_i \gets M(\boldsymbol{x}_i)$
  \EndFor
  \State Apply a base clustering algorithm (e.g., $k$-means) to cluster data points $\boldsymbol{u}_1, \ldots, \boldsymbol{u}_{n+m}$.  The output cluster IDs $\boldsymbol{c}_1, \ldots, \boldsymbol{c}_{n+m}$ for $\boldsymbol{u}_1, \ldots, \boldsymbol{u}_{n+m}$ is that of the data point for $\boldsymbol{x}_1, \ldots, \boldsymbol{x}_{n+m}$
\end{algorithmic}
\end{algorithm}

Regarding the inference or clustering phase in PSC, we aim to efficiently cluster an instance $\boldsymbol{x}_i$, even if $\boldsymbol{x}_i$ does not appear in the initial training data. The model $M$ obtained in the training phase converts each $\boldsymbol{x}_i$ into $\boldsymbol{u}_i$, which serves as an approximation of the $p$ most significant representations of $\boldsymbol{x}_i$. Using this transformation, we can efficiently emulate the computations performed in lines 1 to 4 of Algorithm~\ref{alg:sc}, eliminating the need to construct large similarity and Laplacian matrices and perform computationally expensive matrix operations and eigendecomposition. Consequently, we achieve significant computational savings in time and space but still generate low-dimensional embeddings for the data points $\boldsymbol{x}_i$-s that well preserve the main characteristics.

Once each instance $\boldsymbol{x}_i$ has been transformed into $\boldsymbol{u}_i$, we apply a base clustering algorithm, e.g., $k$-means, on $\boldsymbol{u}_i$-s. The resulting cluster assignments for $\boldsymbol{u}_i$-s are then considered cluster assignments for $\boldsymbol{x}_i$-s. We follow the convention by employing the popular $k$-means algorithm as the clustering method for the low-dimensional data points. However, other base clustering algorithms, e.g., DBSCAN and the Gaussian Mixture Model, can also be applied.

Algorithm~\ref{alg:psc-inf} gives the inference/clustering procedure.

\section{Experiments} \label{sec:exp}

\subsection{Experimental datasets} \label{sec:dataset}

We conducted experiments on both tabular datasets and image datasets. The tabular datasets include the famous Iris, Wine, and BreastCancer datasets. The image datasets include UCIHW, MNIST, and Fashion-MNIST.  %

\subsection{A Comparison of Clustering Quality} \label{sec:cluster-quality-cmp}

\begin{table*}[tb]
\caption{A comparison of the clustering quality of spectral clustering and parametric spectral clustering. We report the mean $\pm$ standard deviation for each score.}
\label{tab:acc-cmp}
\centering
\begin{tabular}{@{}c|ccc|ccc@{}}
\toprule
              & \multicolumn{3}{c|}{SC} & \multicolumn{3}{c}{PSC} \\ \midrule
              & ClusterAcc & ARI & AMI & ClusterAcc  & ARI & AMI \\ \midrule
Iris  & $0.889 \pm 0.00$ & $0.712 \pm 0.00$ & $0.77 \pm 0.00$ & $0.92 \pm 0.025$ & $0.781 \pm 0.06$ & $0.813 \pm 0.038$ \\
Wine  & $0.963 \pm 0.00$ & $0.876 \pm 0.00$ & $0.855 \pm 0.00$ & $0.966 \pm 0.02$ & $0.892 \pm 0.061$ & $0.875 \pm 0.071$ \\
BreastCancer & $0.953 \pm 0.00$ & $0.819 \pm 0.00$ & $0.732 \pm 0.00$ & $0.932 \pm 0.018$ & $0.745 \pm 0.062$ & $0.632 \pm 0.071$ \\
UCIHW & $0.801 \pm 0.067$ & $ 0.743 \pm 0.078$ & $0.858 \pm 0.026$ & $0.825 \pm 0.061$ & $0.764 \pm 0.055$ & $0.855 \pm 0.026$ \\
MNIST & $0.794 \pm 0.04$ & $0.748 \pm 0.03$ & $0.842 \pm 0.013$ & $0.775 \pm 0.046$ & $0.73 \pm 0.049$ & $0.839 \pm 0.013$ \\
Fashion-MNIST & $0.609 \pm 0.015$ & $0.484 \pm 0.009$ & $0.644 \pm 0.009$ & $0.615 \pm 0.041$ & $0.475 \pm 0.028$ & $0.634 \pm 0.009$ \\
\bottomrule
\end{tabular}
\end{table*}

\begin{table}[tb]
\centering
\caption{The network structure of the autoencoder to convert images in MNIST into embeddings ($w$: width of input image; $h$: height of input image).} \label{tab:ae-structure}
\begin{tabular}{@{}lllll@{}}
\toprule
Layer & Type & Input shape & Output shape & Activation \\ \midrule
1     & Flatten & $(w, h)$ & $w \times h$ & ReLU \\
2     & Linear & $w \times h$ & 1568 & ReLU \\
3     & Linear & 1568 & 784 & ReLU \\
4     & Linear & 784 & 392 & ReLU \\
5     & Linear & 392 & 49 & None \\
6     & Linear & 49 & 392 & ReLU \\
7     & Linear & 392 & 784 & ReLU \\
8     & Linear & 784 & 1568 & ReLU \\
9     & Linear & 1568 & $w \times h$ & None \\
10    & Unflatten & $w \times h$ & $(w, h)$ & None \\
\bottomrule
\end{tabular}
\end{table}

\begin{table*}
\centering
\caption{The convolutional neural network structure of the autoencoder to convert images in Fashion-MNIST into embeddings.} \label{tab:iae-structure}
\begin{tabular}{@{}llllllllll@{}}
\toprule
Layer & Type & Input shape & Output shape & Kernel num & Kernel size & Stride & Padding & with BN & Activation \\ \midrule
1     & Conv. & $(w, h, 1)$ & $(w, h, 16)$ & 16 & $5 \times 5$ & 1 & Same & True & ReLU \\
2     & Pooling & $(w, h, 16)$ & $(w/2, h/2, 16)$ & None & $2 \times 2$ & 2 & Valid & False & None \\
3     & Conv. & $(w/2, h/2, 16)$ & $(w/2, h/2, 32)$ & 32 & $3 \times 3$ & 1 & Same & True & ReLU \\
4     & Pooling & $(w/2, h/2, 32)$ & $(w/4, h/4, 32)$ & None & $2 \times 2$ & 2 & Valid & False & None \\
5     & Conv. & $(w/4, h/4, 32)$ & $(w/4, h/4, 1)$ & 1 & $3 \times 3$ & 1 & Same & False & Sigmoid \\
6     & ConvTrans. & $(w/4, h/4, 1)$ & $(w/4, h/4, 32)$ & 32 & $3 \times 3$ & 1 & Same & True & ReLU \\
7     & ConvTrans. & $(w/4, h/4, 32)$ & $(w/2, h/2, 16)$ & 16 & $3 \times 3$ & 2 & Double & True & ReLU \\
8    & ConvTrans. & $(w/2, h/2, 16)$ & $(w, h, 1)$ & 1 & $5 \times 5$ & 2 & Double & False & Sigmoid \\
\bottomrule
\end{tabular}
\end{table*}

We would like to know whether PSC's clustering result is comparable to SC. We compare the SC and PSC's clustering results based on clustering accuracy (ClusterAcc), adjusted rand index (ARI), and adjusted mutual information (AMI).  

Since the experimental datasets contain ground-truth labels, we use these labels to evaluate the quality of a clustering algorithm. However, a clustering algorithm may assign an arbitrary label to each cluster it forms. Therefore, the standard accuracy computation usually underestimates the clustering result. For example, given 5 instances with ground-truth labels as $[1,1,2,2,3]$, if a clustering algorithm assigns their cluster IDs as $[2,2,3,3,1]$, the standard accuracy computation outputs $0 \%$ because none of the predicted cluster IDs matches the ground-truth labels. However, a careful reexamination reveals that the clustering is perfect because all instances of the same ground-truth class belong to the same predicted cluster (and vice versa). Essentially, if two labeled lists are given, and one can become another simply by remapping the label IDs, the cluster assignments of the two lists are identical. In the above example, remapping labels 2 to 1, 3 to 2, and 1 to 3 for the second list makes the two lists identical. As a result, we define the clustering accuracy based on the equation below.

\begin{equation} \label{eq:cluster-acc}
ClusterAcc\left(\boldsymbol{y}, \boldsymbol{\hat{y}}\right) := \max_{\forall {\boldsymbol{\check{y}}} \in P(\boldsymbol{\hat{y}})} \left\{\frac{1}{n}\sum_{i=1}^n I\left(\check{y}^{(i)} = y^{(i)}\right)\right\},
\end{equation}
where $\boldsymbol{y} = [y^{(1)}, \ldots, y^{(n)}]$ is the list of ground-truth labels of size $n$, $\boldsymbol{\hat{y}} = [\hat{y}^{(1)}, \ldots, \hat{y}^{(n)}]$ is a list of predicted cluster IDs, $P(\boldsymbol{\hat{y}})$ returns a set of all identical lists for $\boldsymbol{\hat{y}}$, $I()$ is an indicator function, and $\boldsymbol{\check{y}} = [\check{y}^{(1)}, \ldots, \check{y}^{(n)}]$ is an identical list of $\boldsymbol{\hat{y}}$.  In other words, we compute the accuracies based on the ground-truth label list with all the identical lists of the prediction and return the largest accuracy score.

In addition to clustering accuracy, we also use ARI and AMI to evaluate the quality of the clustering results. ARI and AMI are biased towards different clustering assignments: ARI likes a balanced partition~\cite{romano2016adjusting}, and AMI favors an unbalanced partition. We report both to fairly assess different situations.

Since both SC and PSC are nondeterministic, executing either multiple times on a consistent dataset produces varied clustering outcomes. To fairly evaluate each method, we conduct each experiment 5 times and subsequently present the mean and standard deviation derived from these 5 trials. 

Table~\ref{tab:acc-cmp} gives the three quality measurements for the SC and PSC clustering results in the six datasets. SC and PSC yield comparable results on all datasets. Interestingly, PSC sometimes outperforms SC even though PSC approximates the low-dimensional representations computed by SC.  This is probably because the learning process implicitly introduces regularization factors so that PSC learns a better mapping.

Raw images are often characterized by complex structures, which hinders the effectiveness of traditional clustering algorithms. Therefore, we applied an autoencoder as a preprocessing step to extract the essential information of each image in the MNIST dataset into a compact vector that captures the critical patterns in an image. The structure of the autoencoder is shown in Table~\ref{tab:ae-structure}. We use the bottleneck layer (the output of layer 5 in Table~\ref{tab:ae-structure}) as the compact representation of the input image. Fashion-MNIST images are more complicated than MNIST, so we use an image autoencoder to convert images into vectors so that the spatial structures are reserved. Table~\ref{tab:iae-structure} gives the structure of the image autoencoder. We use the output of layer 7 as the extracted features.

Regarding the hyperparameters, we define $n_1, n_2, n_3$ (as specified in Table~\ref{tab:nn}) as follows: 32, 64, 32 for the Iris dataset; 26, 52, 26 for the Wine dataset; 60, 120, 60 for the BreastCancer dataset;  196, 392, 196 for both the MNIST dataset and the Fashion-MNIST.  While further fine-tuning these hyperparameters may yield better clustering results, we did not spend much effort tuning them, as our primary focus is to show that PSC can cluster big data.

\subsection{Training Data Size, Clustering Quality, Empirical Execution Time, and Empirical Memory Usage}

\begin{table*}[tb]
\caption{A comparison of SC and PSC with different sampling ratios in terms of the execution time (seconds), peak memory usage (MB), and clustering quality on the MNIST dataset.}
\label{tab:sc-psc-train-size-quality-time-cmp}
\centering
\begin{tabular}{@{}cccccc@{}}
\toprule
Method & Execution time (s) & Peak memory usage (MB) & ClusterAcc     & ARI            & AMI            \\ \midrule
SC & 2462 & 5331 & $0.794 \pm 0.04$ & $0.748 \pm 0.03$ & $0.842 \pm 0.014$ \\ \midrule
PSC $(r=1/6)$  & \makecell[l]{ Training: $453$ ($\color{red} \downarrow 82\%$) \\Inference: $0.443 \pm 0.027$ } & \makecell[l]{Training: $1032$ ($\color{red} \downarrow 81\%$) \\Inference: $92.83 \pm 1.965$ } & $0.732 \pm 0.076$ & $0.684 \pm 0.068$ & $0.792 \pm 0.03$ \\ \midrule
PSC $(r=2/6)$  & \makecell[l]{ Training: $533$ ($\color{red} \downarrow 78\%$)\\Inference: $0.419 \pm 0.088$  } & \makecell[l]{Training: $1717$ ($\color{red} \downarrow 68\%$)
\\Inference: $92.63 \pm 2.09$ } & $0.739 \pm 0.089$ & $0.68 \pm 0.066$ & $0.798 \pm 0.02$ \\  \midrule
PSC $(r=3/6)$  & \makecell[l]{ Training: $706$ ($\color{red} \downarrow 71\%$)
\\Inference: $0.437 \pm 0.048$ } & \makecell[l]{Training: $2532$ ($\color{red} \downarrow 53\%$) \\Inference: $96.21 \pm 1.785$ } & $0.764 \pm 0.041$ & $0.712 \pm 0.044$ & $0.825 \pm 0.016$ \\  \midrule
PSC $(r=4/6)$  & \makecell[l]{ Training: $1029$ ($\color{red} \downarrow 58\%$)
\\Inference: $0.379 \pm 0.06$ } & \makecell[l]{Training: $3328$ ($\color{red} \downarrow 38\%$) \\Inference: $96.54 \pm 2.271$ } & $0.775 \pm 0.046$ & $0.73 \pm 0.049$ & $0.839 \pm 0.013$ \\  \midrule
PSC $(r=5/6)$  & \makecell[l]{ Training: $1472$ ($\color{red} \downarrow 40\%$) \\Inference: $0.491 \pm 0.036$ } & \makecell[l]{Training: $4937$ ($\color{red} \downarrow 7\%$)
\\Inference: $96.5 \pm 2.15$ } & $0.819 \pm 0.039$ & $0.753 \pm 0.023$ & $0.84 \pm 0.01$ \\
\bottomrule
\end{tabular}
\end{table*}

In Section~\ref{sec:cluster-quality-cmp}, we have demonstrated the quality of SC and PSC clustering using identical training instances. However, because PSC utilizes a parametric model to map data points to lower-dimensional representations, there exists an opportunity to use a subset of the provided data points as training data for PSC. Here, we investigate the efficiency and effectiveness of PSC utilizing a subset of training instances.

In particular, this section compares execution time, peak memory consumption, and clustering quality between SC and PSC, considering the varying rates of training data sampling for PSC. We use the MNIST dataset in this section for experiments. Note that applying spectral clustering on the MNIST dataset using a modern laptop is sometimes infeasible due to the substantial memory requirements of the modern spectral clustering algorithm. For example, if we want to cluster the entire MNIST dataset (which contains $70,000$ samples) using spectral clustering, a na\"{i}ve implementation requires creating large matrices (e.g., $\boldsymbol{D}$ and $\boldsymbol{L}$ in Algorithm~\ref{alg:sc}), each of them with shape $70,000 \times 70,000$. If each entry is a single-precision floating point (4 bytes), a single matrix needs $4~\textrm{bytes} \times 70,000^2 \approx 19.6~\textrm{GB}$.

We empirically explore the influence of different training data sampling rates on the clustering quality, execution time, and memory utilization of parametric spectral clustering when applied to the MNIST dataset. Table~\ref{tab:sc-psc-train-size-quality-time-cmp} describes the experimental results. First, the training and inference duration of the PSC is considerably faster than the execution time of the SC. The PSC's training process requires only a portion of the training data, which is why it is faster than SC. The inference time for PSC is almost negligible. Second, for the same reason, the peak memory utilization during the PSC's training and inference stages remains below the SC's peak memory requirement. The peak memory consumption is approximated using the memory profiler \texttt{mprof}.\footnote{\url{https://pypi.org/project/memory-profiler/}} Third, we study the connection between the training data sampling rate and the clustering quality achieved by the PSC. The results demonstrate that increasing the sampling rate can lead to an improvement in clustering quality. However, the enhancement is not considerable, and the gap becomes insignificant when the sampling rate exceeds a certain threshold. In essence, after reaching this threshold, the incremental benefit arising from including additional training instances becomes negligible and can significantly increase the cost of training. Consequently, we conclude that the PSC generates satisfactory clustering results using more economical training resources without sacrificing substantial clustering quality.

\subsection{Quality and Computational Cost of Incremental Clustering}

\begin{table*}[tb]
\caption{The quality and computation cost of incremental clustering when the number of extra instances varies}
\label{tab:incremental-clustering-cmp}
\centering
    \begin{tabular}{@{}cccccc@{}}
    \toprule
    Size of extra instances & Clustering time (s) & Peak memory usage (MB) & ClusterAcc & ARI  & AMI  \\ \midrule
    2,000   & $0.569 \pm 0.012$ & $239.8 \pm 2.949$ & $0.738 \pm 0.045$ & $0.675 \pm 0.054$ & $0.811 \pm 0.022$ \\
    4,000   & $0.59 \pm 0.026$ & $198.6 \pm 3.66$ & $0.74 \pm 0.042$ & $0.705 \pm 0.022$ & $0.82 \pm 0.01$ \\
    6,000   & $0.634 \pm 0.022$ & $207.9 \pm 3.145$ & $0.745 \pm 0.071$ & $0.673 \pm 0.099$ & $0.816 \pm 0.032$ \\
    8,000   & $0.681 \pm 0.095$ & $262.4 \pm 4.366$ & $0.744 \pm 0.052$ & $0.705 \pm 0.039$ & $0.826 \pm 0.018$ \\
    10,000  & $0.659 \pm 0.058$ & $272.7 \pm 3.838$ & $0.795 \pm 0.057$ & $0.739 \pm 0.072$ & $0.844 \pm 0.026$ \\ \bottomrule
    \end{tabular}
\end{table*}

This section demonstrates the efficacy of PSC in handling incremental clustering tasks. 

We use a subset of $20,000$ instances from MNIST as initial training data to train the PSC model. Subsequently, we simulate an incremental clustering scenario by introducing incremental batches of new instances that require clustering. Specifically, we consider scenarios with additional samples of sizes $2,000$, $4,000$, $6,000$, $8,000$, and $10,000$. PSC is requested to cluster the initial $20,000$ instances and the extra samples using the model trained only on the initial $20,000$ training instances, thus fulfilling the incremental clustering.

The results of these incremental clustering simulations are presented in Table~\ref{tab:incremental-clustering-cmp}. As the number of newly arrived instances increases, there are slight changes in both the clustering time and the peak memory usage. This result validates that the PSC exhibits remarkable efficiency for incremental clustering tasks. Meanwhile, including more extra instances seems to improve the clustering quality marginally. We suspect that more instances may reveal a clearer pattern among different groups, thereby increasing the quality of the clustering results. On the contrary, SC is not able to manage incremental clustering. Therefore, even adding a small number of data points requires complete retraining, which is time consuming and memory intensive.

\section{Discussion} \label{sec:disc}

In this paper, we have proposed an innovative approach called parametric spectral clustering to address the limitations of traditional spectral clustering in the handling of large datasets and the support of incremental clustering. By leveraging a combination of training with a subset of instances and efficient low-dimensional projection, the PSC achieves efficient computation and economical memory usage while maintaining competitive clustering accuracy. Furthermore, we have demonstrated the effectiveness of PSC through extensive experiments on various datasets. Our results show that the PSC outperforms traditional spectral clustering regarding computational efficiency and peak memory usage. Additionally, we have validated the applicability of PSC in online clustering scenarios, where new data points can be clustered on the fly without retraining the entire model. Overall, our contribution extends the application of spectral clustering to domains with big data and real-time clustering requirements, thus opening up new possibilities for efficient and effective nonlinear clustering in various fields.

As an ongoing work, we are developing a user-friendly library that provides a scikit-learn-like interface for PSC. Users can use PSC with the commands \texttt{psc.fit(X)} and \texttt{psc.predict(X)} to cluster the data points. This library will simplify the implementation and usage of PSC, offering seamless integration with existing machine learning workflows. By providing accessible tools and resources, we aim to empower a broader range of users, including practitioners and researchers with varying levels of expertise in machine learning and data analysis, to explore the potential of PSC in their projects. This will facilitate greater adoption and further advances in the field, fostering collaboration and knowledge exchange among the research community.

A potential issue with the PSC is its inefficiency in adapting to data drift. The computation of a new instance's ground-truth embedding through eigendecomposition in spectral clustering makes the acquisition of ground-truth labels for training targets resource-intensive. Therefore, developing methods to efficiently manage data drift is a direction we are eager to explore.

\bibliographystyle{unsrt}  
\bibliography{ref}

\end{document}